\documentclass[conference]{IEEEtran}
\usepackage{times}
\usepackage{authblk}
\usepackage{graphicx}
\usepackage{amsmath}
\usepackage{hyperref}
\usepackage{color}
\usepackage{comment}

\title{A Technique for Isolating Lexically-Independent Phonetic Dependencies in Generative CNNs}

\author{Bruno Ferenc \v{S}egedin\\
\texttt{bruno\_ferenc\_segedin@brown.edu}}
\affil{Cognitive \& Psychological Sciences; Program in Linguistics, Brown University}

\date{}

\begin{document}

\maketitle

\begin{abstract}
The ability of deep neural networks (DNNs) to represent phonotactic generalizations derived from lexical learning remains an open question. This study (1) investigates the lexically-invariant generalization capacity of generative convolutional neural networks (CNNs) trained on raw audio waveforms of lexical items and (2) explores the consequences of shrinking the fully-connected layer (FC) bottleneck from 1024 channels to 8 before training. Ultimately, a novel technique for probing a model's lexically-independent generalizations is proposed that works only under the narrow FC bottleneck: generating audio outputs by bypassing the FC and inputting randomized feature maps into the convolutional block. These outputs are equally biased by a phonotactic restriction in training as are outputs generated with the FC. This result shows that the convolutional layers can dynamically generalize phonetic dependencies beyond lexically-constrained configurations learned by the FC.
\end{abstract}

\section{Introduction}
A goal of modeling speech acquisition with deep neural networks (DNNs) is to build models whose internal states meaningfully correspond to human-like processes or representations (\cite{abdullah2021familiar}, \cite{baas2024disentanglement}). The current paper is concerned with DNNs' capacity to represent phonetic or phonotactic generalizations derived from lexical learning. Focusing on generative CNNs, this paper presents a technique for probing learning of lexically-independent phonetic dependencies in models trained on audio waveforms of lexical items. This technique produces lexically unconstrained sequences of linguistically interpretable audio outputs that can be used to infer the model's lexically independent generalization. 

\subsection{Sound patterns and the Lexicon}
Phonetics and phonotactics can operate largely independently of the lexicon (\cite{davidson2011phonetic}, \cite{hayes2008maximum}). An example that highlights this intuition is the fact that a non-word utterance like "blickly" can be judged as adhering to the phonology of English, even though it is not part of the lexicon. The relative autonomy of phonetics and the lexicon is exemplified in heavily ``foreign-accented" speech, wherein L2 words might be adapted to adhere to the phonotactics of an L1 language (e.g. \cite{davidson2011phonetic}).

Phonetic and phonotactic restrictions on lexical forms are governed by a distinct set of computational constraints compared to those shaping the composition of lexical forms within the space of phonotactically well-formed words. Phonetics and phonotactics are often described as operating over local representations like adjacent sounds (e.g. \cite{kawahara2004locality}, \cite{goldsmith1990autosegmental}), and being grounded in articulatory phenomena that can often be constrained by the inherent locality of interactions between certain articulatory gestures (e.g. \cite{hayes1999phonetically}, \cite{pierrehumbert2000phonetic}). In contrast, the factors that determine the attestation of some phonetically or phonotactically well-formed lexical forms over others also include pressures for communicative disambiguation (\cite{trott2020human}, \cite{piantadosi2012communicative}) that are not locally constrained. It is thus desirable for a computational model to be able to represent local phonetic/phonotactic dependencies separately from the lexical configurations in which they occur. 

\subsection{Modeling phonotactics from raw speech}
Attempts to model the knowledge that underlies phonetic dependencies with DNNs have generally relied either on discrete symbolic representations as training input or on supervised learning and explicit linguistic labels (e.g. \cite{mayer2020phonotactic}, \cite{stoianov2021exploring}). Recently, latent space models like auto-encoders and generative adversarial networks (GANs) have been touted as being capable of learning categorical phonological phenomena from raw speech input in an unsupervised manner and without explicit labels (e.g. \cite{beguvs2020modeing}, \cite{chen2023exploring}). For example, Shain \& Elsner \cite{shain2019measuring}, argue that an auto-encoder trained to reproduce individual phonemes without any explicit labels contains evidence of phoneme-general features in its latent space representations. 

More recent work using GANs trained on waveforms of multi-segment utterances like lexical items has shown that the generator's latent space can map a specific segment category to a single latent variable (\cite{beguvs2020modeing}; \cite{chen2023exploring}; \cite{beguvs2021ciwgan}). For example, \cite{beguvs2020modeing} showed that single latent variables in a WaveGAN model (\cite{donahue2018adversarial})  are capable of uniquely modulating the manifestation of certain phonemic categories like /s/ and /t/ in the output. Chen \& Elsner \cite{chen2023exploring} test the ability of GANs to approximate learning of phonological features and find that the concurrent adjustment of two latent variables predictably modulates the degree of nasality on a vowel in the output. Some of this work has more directly tested how these models generalize phonotactic-like dependencies from lexical inputs (e.g. \cite{beguvs2020modeing}, \cite{barman2024unsupervised}). This work leverages segment-specific latent variables to probe models' phonotactic-like generalizations about which sounds can co-occur in an output. For example, Begu\v{s} \cite{beguvs2021identity} showed that the model generalizes a reduplication operation to a sound for which there was no reduplication in training- by concurrently manipulating the latent variable correlated with that sound and that correlated with the presence of reduplication. 

\subsection{Fully-connected \& convolutional layers}
A challenge for the interpretability of the aforementioned latent space approaches when it comes to modeling phonetic dependencies separately from lexical knowledge is that the generative models are trained to exclusively produce outputs that share the lexical configuration of their training data (e.g. \cite{barman2024unsupervised},  \cite{chen2023exploring}). Lexically constrained outputs are learned in the fully-connected layer (FC), where latent variables manipulate variable-specific weight matrices that are structured along the time dimension and thus fix sounds within specific lexical templates (e.g. \cite{vsegedin2025exploring}). Thus, while latent space interpolation is invaluable as a diagnostic for detecting disentangled lexical and sublexical representations, it is limited in its utility for exhaustively exploring a model's lexically-independent learning, because it manipulates lexically-constrained representations in the FC. 

Convolutional layers, in contrast, are translation-invariant (e.g. \cite{kauderer2017quantifying}). Unlike FC weights, convolutional kernels are not bound to particular temporal positions in the input, so they produce waveform outputs that preserve the basic temporal structure of whatever feature map input they receive. Convolutional layers are thus potentially well suited for capturing lexically invariant generalization in generative CNNs. 

\section{Current Goals}
\subsection{A technique for isolating phonetic dependencies in CNNs}
\begin{figure}
    \centering
    \includegraphics[width=0.75\linewidth]{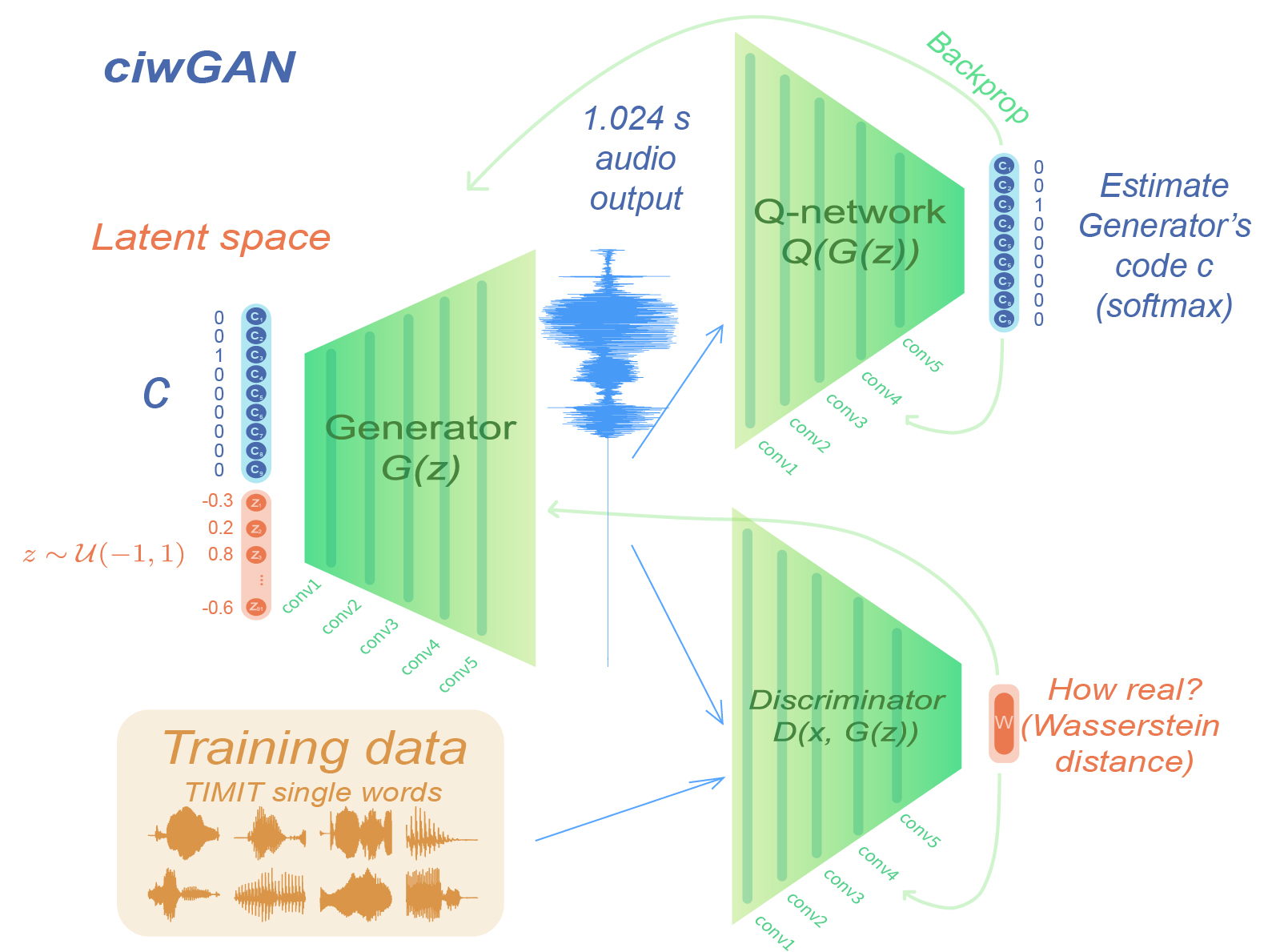}
    \caption{Schematic of ciwGAN architecture, from \cite{beguvs2023basic}.  The current study does not use TIMIT words.}
    \label{fig:ciwGAN}
\end{figure}

This paper proposes and tests a technique that leverages the translation-invariance of convolutional layers to exhaustively probe the lexically-independent generalization ability of generative CNNs like the WaveGAN generator. This technique consists of inputting fully random feature maps into the convolutional stack of a trained model instead of the feature maps produced by the FC. Waveform outputs produced this way can in principle reveal a model's lexically-invariant learning in an empirically measurable way. 
The proposed technique is tested using ciwGAN (see \cite{beguvs2021ciwgan}, Fig. \ref{fig:ciwGAN}), an innovation of the WaveGAN architecture \cite{donahue2018adversarial}, and described in more detail in the next section. This paper is primarily concerned with the generator's generative CNN architecture, and leverages the ciwGAN training objective as a proven method for yielding interpretable lexical outputs. 
Fig. \ref{fig:generator} sketches the basic generator model used in the current study. The conventional means of sampling outputs from the generator is to input uniformly distributed latent variables between -1 and 1, (plus one-hot vectors for a subset of features in the case of ciwGAN) to the FC which in turn produces the feature map that ultimately gets sequentially up-sampled to an audio waveform via the convolutional stack.
This paper propose sampling outputs of a trained model by replacing the feature maps produced by the FC in Fig. \ref{fig:generator} with equally-sized feature maps composed of values sampled randomly from a uniform distribution. Given the translation-invariance of convolutional layers, outputs generated from such random feature maps should, in principle, carry measurable information about what linguistic generalizations the generator can make independent of the lexical configurations enforced in the FC. Comparing these outputs to those generated with FC feature maps could thus be useful for inferring what aspects of generated outputs are determined uniquely by the lexically-constrained FC.    

\subsection{Introducing a bottleneck to the FC}
\begin{figure}
    \centering
    \includegraphics[width=0.75\linewidth]{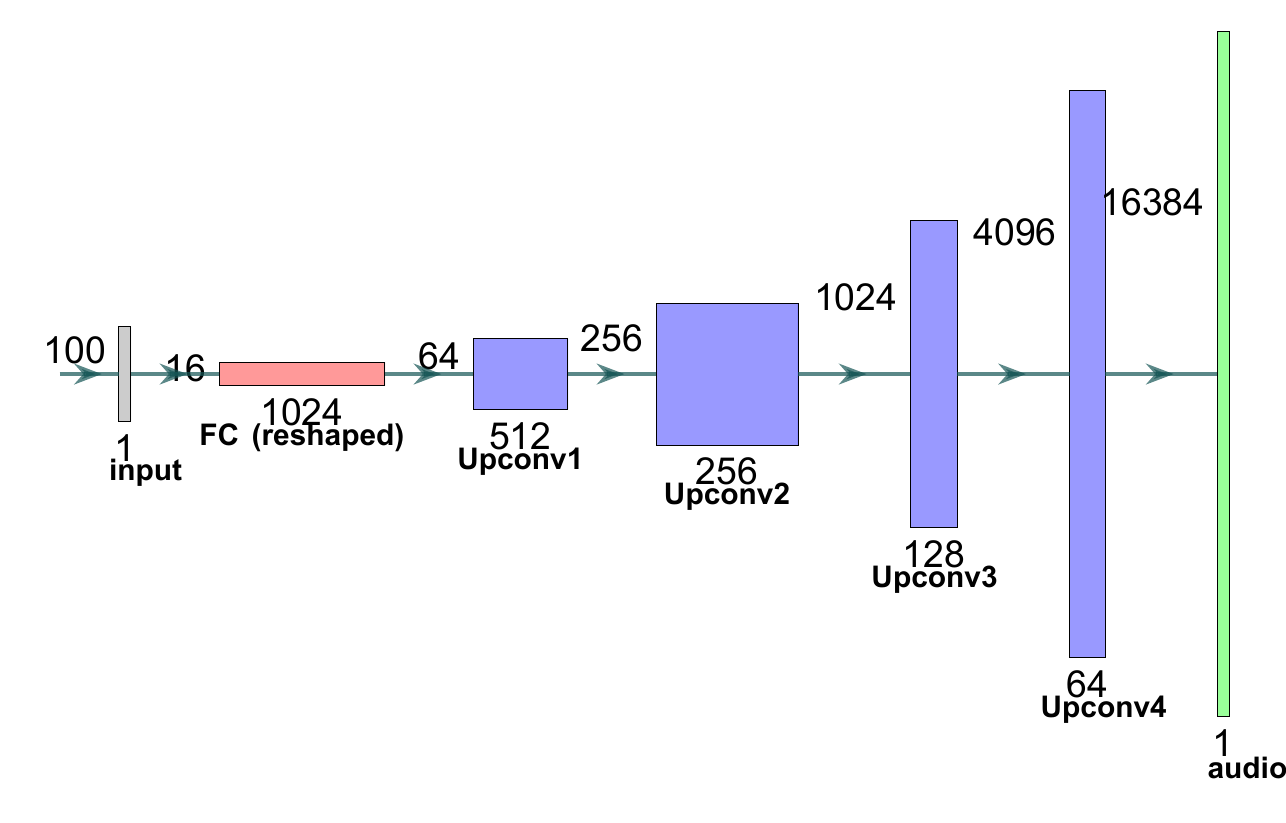}
    \caption{Generator architecture in WaveGAN-based models, showing the dimensionality of the output of each layer. This paper proposes changing the FC output from 1024x16 (represented in red), to as small as 8x16.}
    \label{fig:generator}
\end{figure}

The utility of the above technique to linguistic and cognitive modeling hinges on the resulting waveform outputs being variable and containing interpretable linguistic (phonetic) structure. Past studies using WaveGAN for lexical learning have stated that a certain proportion of outputs generated conventionally (i.e. inputting latent space values within training range) are noisy and linguistically-uninterpretable (e.g. \cite{begusCSL}). Thus, inputting entirely random feature maps into the convolutional layers may yield entirely uninterpretable outputs. To address this issue, this paper proposes training the generator with a drastically reduced number of channels in the FC: replacing the 1024x16 feature map in Fig. \ref{fig:generator} with an 8x16 feature map. If the small-bottleneck version of the model is still successful in reproducing outputs resembling training data, then outputs generated via random 8x16 feature maps may be more phonetically structured (i.e. contain sequences of recognizable and distinct phoneme-like sounds) than those produced with random 1024x16 feature maps in the 1024ch model. Specifically, the space of possible 1024x16 feature maps is likely to be sparsely sampled by the FC during training, and thus a drastically reduced space of possible feature maps may make it likelier that the randomly generated maps will resemble those encountered by the convolutional stack during training. This manipulation is also motivated by recent work suggesting that information is redundantly encoded across channels in the FC feature map (\cite{vsegedin2025exploring}), and it is independently motivated given the practical advantages of reducing the amount of model parameters. 

\section{Experiment}
The goal of the current experiment is to (1) test whether a convolutional block that gets a random feature-map as input yields linguistically interpretable outputs, (2) explore how significantly shrinking the FC mediates the linguistic interpretability of these outputs and (3) test whether these outputs align with a particular local phonotactic-like restriction for which there is evidence in the training data even when the FC is bypassed. 

\subsection{Data}
Two sets of models are each trained on two distinct sets of training data. The first training set consists of 1800 total tokens, 300 tokens of each of the following words, generated using speecht5 text-to-speech model (\cite{ao-etal-2022-speecht5}): ['scope', 'gasp', 'expect', 'sticks', 'whisky', 'past']. The lexical types are chosen to license the generalization that /s/ never occurs directly before a vowel (referred to here as an "SV transition"). Another set of models is trained on a second training set where three of the original six words are replaced with words that do have SV transitions: ['scope', 'gasp', 'expect', 'bo\textit{ssy}', '\textit{sa}p', '\textit{si}x']. Each lexical type has 300 acoustic tokens, which are generated by speecht5 using 300 speaker embeddings from the same male English speaker.

\subsection{Model \& Training}
To optimize the generator, the ciwGAN training objective is used (\cite{beguvs2021ciwgan}, see Fig. \ref{fig:ciwGAN}). Under this framework (following \cite{beguvs2021ciwgan}), the generator is optimized to maximize the discriminator score \textit{D(G(z))}, while the discriminator's goal is to maximize the Wasserstein distance between scores of real training data \textit{D(x)} and \textit{D(G(z)).} Both the Q-network and generator maximize \textit{Q(G(z))}; the Q-network takes as input a generated sound and is optimized to classify the one-hot code that initialized the generator, which is in turn optimized to facilitate this classification and thus generate informative outputs. 

Two conventional ciwGAN models are trained with 6 one-hot codes to match the 6 lexical types in the training data. Two equivalent models with the generator's FC shrunk to 8 channels instead of 1024 are also trained. All other model specifications follow those for ciwGAN in \cite{beguvs2021ciwgan}. All models are trained on a high-performance computing cluster at Brown University's Center for Visualization and Computation. Model training was performed on a single NVIDIA GPU for each model with a total runtime of approximately 5 hours per session. 

\subsection{Analysis}
Outputs for both models are analyzed after 1350 epochs (33,600 steps) of training. 200 waveform outputs are generated for each combination of bottleneck size (1024- and 8-channel) and each output-generation technique (FC-generated using latent variables, or Conv-only, using random feature maps). Linguistically interpretable outputs are annotated using Praat software (\cite{praat}). The author manually annotated all outputs, marking intervals that measure the distance between an abrupt drop in /s/ amplitude (high frequency noise) and the first glottal pulse of a voiced vowel. The dependent variable is voice onset time (VOT; relative to /s/ completion, not stop release burst as it is conventionally measured). Longer average /s/-vowel VOTs can be interpreted as evidence of a bias against /s/ immediately preceding a vowel. 

\subsection{Results}
\begin{figure}
    \centering
    \includegraphics[width=0.75\linewidth]{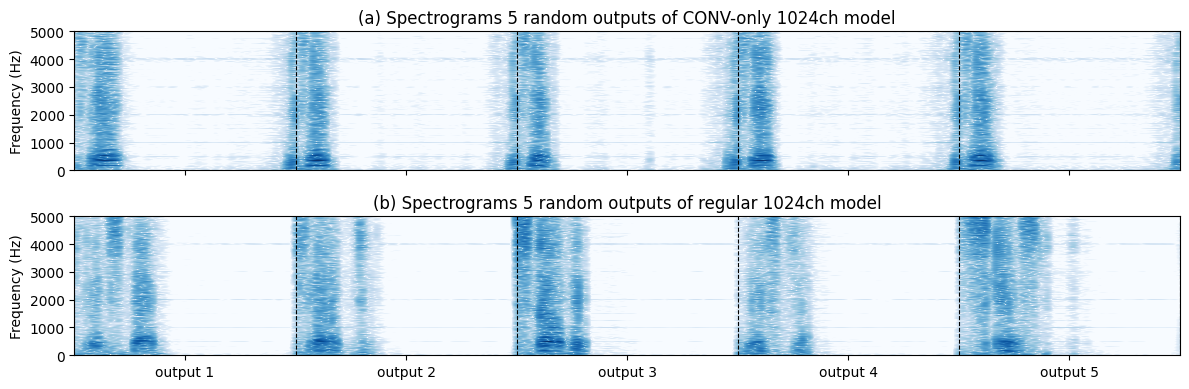}
    \caption{1024ch models' Spectrograms of Conv-only waveform outputs show little variability and almost no linguistic information. The spectral information near the beginning and end of the signal are likely due to boundary effects or artifacts of zero-padding.}
    \label{fig:1024ch_spectrograms}
\end{figure}

\begin{figure}
    \centering
    \includegraphics[width=0.75\linewidth]{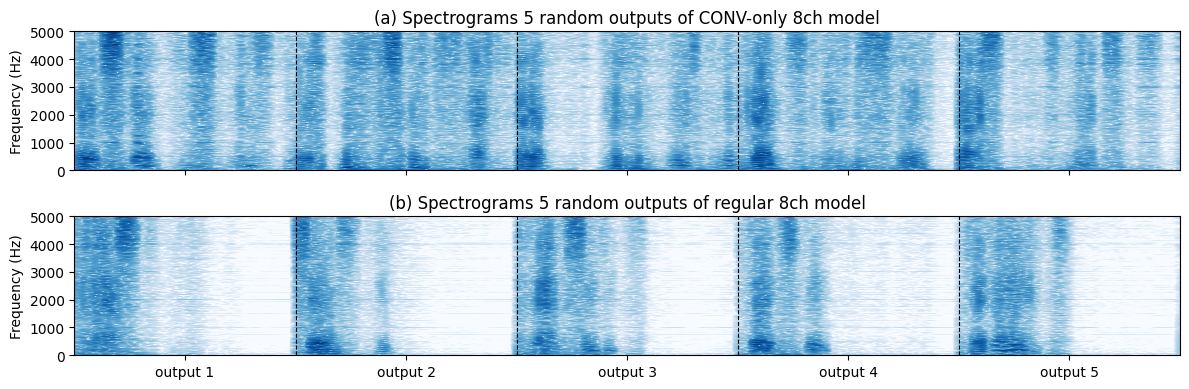}
    \caption{The CONV-only outputs of the 8ch model yield phonotactic-like linguistically structured variability throughout the signal. The local formant structure in the FC outputs (most clearly /s/ and vowel formants) are also represented in the CONV-only outputs. }
    \label{fig:8ch_spectrograms}
\end{figure}

\begin{figure}
    \centering
    \includegraphics[width=0.75\linewidth]{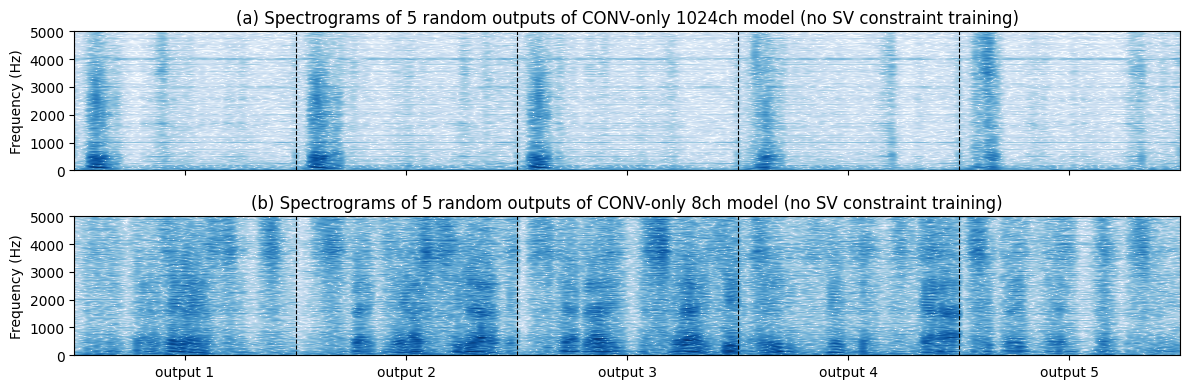}
    \caption{Direct comparison of 8ch and 1024ch models' spectrograms for Conv-only conditions shows that the 1024ch model does not yield informative generalizations.}
    \label{fig:8ch_1024ch}
\end{figure}

There is a qualitative difference between the 1024-channel model and 8-channel models in their Conv-only (random feature map) outputs. In the SV constraint condition, both bottleneck conditions yield linguistically interpretable outputs generated via latent space inputs to the FC. However, among the Conv-only models' outputs, the 8-channel model exhibits highly variable and spectrally structured waveforms along the entire output, while the 1024-channel model does not (Figs. \ref{fig:1024ch_spectrograms}, \ref{fig:8ch_spectrograms}). This qualitative discrepancy between 8ch and 1024ch outputs is replicated when the models are trained on the dataset without an SV restriction (see Fig.\ref{fig:8ch_1024ch}).

\begin{figure}
    \centering
    \includegraphics[width=0.75\linewidth]{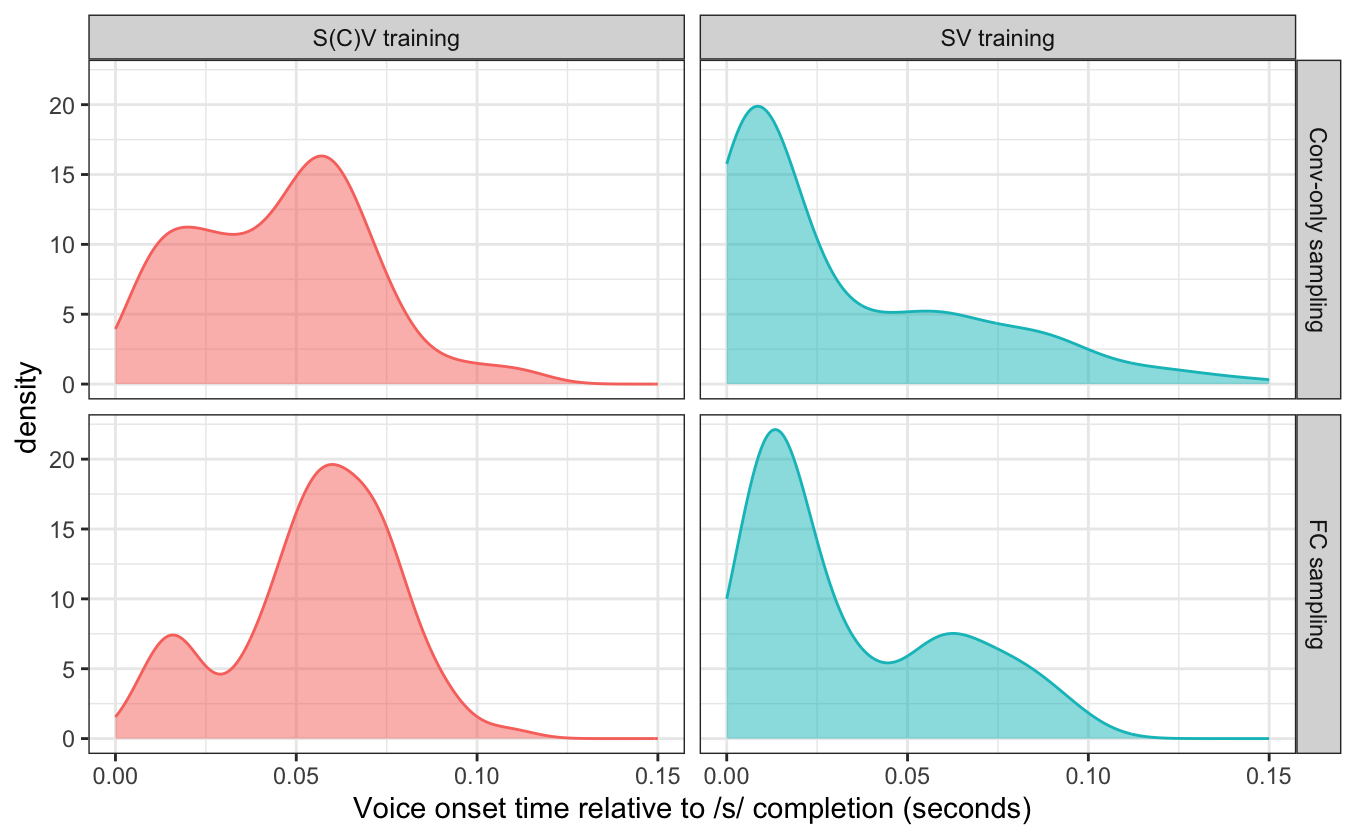}
    \caption{S-V voice onset time for sampled outputs of four 8ch models}
    \label{fig:VOT_final_fig}
\end{figure}

Next, how is learning of the training pattern mediated by the exclusion of the FC layer? This analysis is restricted to 8ch outputs, as 1024ch outputs in the Conv-only condition do not preserve any linguistic information. The qualitative patterns of VOT are consistent with the training condition regardless of whether or not the FC was bypassed (Fig. \ref{fig:VOT_final_fig}). Outputs of the model trained on lexical items with no restriction against SV transitions appear to have shorter VOTs.

To determine whether Conv-only outputs deviate from the FC-constrained outputs in their learning of the SV restriction, the following linear regression model was fit using glm in R (\cite{R-base}), where each predictor is a binary variable: \textit{VOT $\sim$ training\_condition + conv\_only + training\_condition:conv\_only}. 

\textit{Training\_condition} refers to whether or not the training data were consistent with the restriction against SV transitions. \textit{Conv\_only} refers to whether or not the outputs were generated with random feature maps (as opposed to feature maps from the FC layer).  

\textit{Training\_condition} significantly predicted \textit{VOT}  (estimate = 0.021, SE = 0.004, t = 5.12, p $<$ 2e-16) suggesting that outputs were biased by the training pattern. There was no main effect of \textit{conv\_only} (estimate = -0.0003, SE = 0.004, t = -0.08, p = 0.939), suggesting that bypassing the FC did not introduce an inherent bias for producing long or short VOTs. Critically, there was no interaction between \textit{conv\_only} and \textit{training\_condition} (estimate = -0.0076, SE = 0.006, t = -1.35, p = 0.177); there is thus no evidence of the FC being directly involved in enforcing adherence to the constraint from the training data. Rather, this result is consistent with the notion that the convolutional layers account for the models' adherence to this particular phonotactic pattern in training data.
\section{Discussion}
This study's results support the proposal that convolutional layers of generative CNNs can uniquely represent local dependencies generalized from lexical training data, independently of the FC layer. In models with narrow FC bottlenecks, bypassing the FC and generating waveform outputs by inputting random feature maps into the convolutional stack yields observable evidence of what local phonetic dependencies are generalized beyond lexical configurations enforced by the FC. This technique can be loosely thought of as an acoustic version of n-phone models that generate strings of phoneme symbols based on the symbols' local dependencies inferred from their distributions in lexical items (e.g. \cite{trott2020human}, \cite{dautriche2017words}). This approach should complement existing methods for linguistic interpretability of latent space models that establish phonetically-invariant categories or features (\cite{beguvs2020modeing}, \cite{chen2023exploring}).  

This study shows that ciwGAN generators can produce interpretable outputs even when trained with fewer than 1\% of the original number of parameters in the FC. Beyond its advantages for interpretability, a generator with a narrow FC bottleneck is arguably more plausible as a cognitive model than the standard 1024-channel bottleneck used in WaveGAN and ciwGAN. Under the small bottleneck, the FC's trained parameters constrain the output to a subspace of lexical items from a larger space of already phonetically interpretable waveforms (i.e. those sampled directly from the convolutional stack in the 8ch models, see Fig. \ref{fig:8ch_1024ch}). Lexical selection in human speech production similarly involves accessing and manipulating relatively coarse sublexical representations within a narrow set of dimensions rather than selecting phonetically precise outputs from a large space of mostly noisy acoustic signals-- something the FC weights are tasked with in the 1024ch model. Future work should evaluate how this dimensionality reduction affects other desiderata of generative CNNs, like latent space disentanglement (e.g. \cite{beguvs2020modeing}). 

\section{Conclusion}
This paper provides empirical support for the following claims: (1) WaveGAN-based generators can yield interpretable and variable acoustic outputs even when trained with fewer than 1\% of parameters in the FC, (2) The convolutional block of small-bottleneck models can generate interpretable outputs while bypassing the FC with random feature map inputs (something the original 1024-channel model is incapable of), and (3) These outputs show some evidence of phonotactic-like generalization: their local phonetic dependencies are consistent with lexical patterns produced by the FC and with evidence for a phonotactic constraint in the training data. This paper ultimately contributes to the interpretability of CNNs for phonological learning by showing how the division between the FC and convolutional layers' computational constraints parallel those between lexical knowledge and phonetic dependencies respectively.   

\section{Acknowledgments}
I thank Ga\v{s}per Begu\v{s} and members of the UC Berkely Speech \& Computation lab for helpful discussions.

\label{section:references}

\bibliographystyle{IEEEtran}
\bibliography{mybib}

\end{document}